\documentclass[10pt,twocolumn,letterpaper]{article}

\usepackage{cvpr}              % To produce the CAMERA-READY version 终版论文
\definecolor{cvprblue}{rgb}{0.21,0.49,0.74} 
\usepackage[pagebackref,breaklinks,colorlinks,allcolors=cvprblue]{hyperref}

\def\paperID{*****} % *** Enter the Paper ID here 论文的ID。在最终版本提交时，用户需要用实际的论文ID替换 *****。
\def\confName{CVPR}
\def\confYear{2025}

\title{Mamba-VA: A Mamba-based Approach for Continuous Emotion Recognition in Valence-Arousal Space}

\author{
Yuheng Liang, Feng Liu, Mingzhou Liu, Yu Yao\\
Nanjing University of Posts and Telecommunications\\
Jiangsu Key Laboratory of Intelligent Information Processing and Communication Technology\\
{\tt\small \{1023010418,liuf,1023162807,1023010419\}@njupt.edu.cn}
\and
Zheyu Wang\\
Nanjing University of Science and Technology ZiJin College\\
School of Computer and Artificial Intelligence\\
{\tt\small wangzheyu@njust.edu.cn}
}

\begin{document}
\maketitle
\begin{abstract}
Continuous Emotion Recognition (CER) plays a crucial role in intelligent human-computer interaction, mental health monitoring, and autonomous driving. Emotion modeling based on the Valence-Arousal (VA) space enables a more nuanced representation of emotional states. However, existing methods still face challenges in handling long-term dependencies and capturing complex temporal dynamics. To address these issues, this paper proposes a novel emotion recognition model, Mamba-VA, which leverages the Mamba architecture to efficiently model sequential emotional variations in video frames.
First, the model employs a Masked Autoencoder (MAE) to extract deep visual features from video frames, enhancing the robustness of temporal information. Then, a Temporal Convolutional Network (TCN) is utilized for temporal modeling to capture local temporal dependencies. Subsequently, Mamba is applied for long-sequence modeling, enabling the learning of global emotional trends. Finally, a fully connected (FC) layer performs regression analysis to predict continuous valence and arousal values.
Experimental results on the Valence-Arousal (VA) Estimation task of the 8th competition on Affective Behavior Analysis in-the-wild (ABAW) demonstrate that the proposed model achieves valence and arousal scores of 0.5362 (0.5036) and 0.4310 (0.4119) on the validation (test) set, respectively, outperforming the baseline. The source code is available on GitHub: \url{https://github.com/FreedomPuppy77/Charon.}
\end{abstract}    
\section{Introduction}
\label{sec:intro}

Affective Computing, an essential research direction in computer vision and human-computer interaction (HCI), aims to equip computers with the ability to understand and recognize human emotions~\cite{Picard97}. Among various emotion recognition tasks, video-based emotion recognition has gained significant attention in recent years due to its non-intrusive nature and the richness of visual information~\cite{Poria17}.

Traditional emotion recognition methods predominantly rely on discrete emotion classification (e.g., happiness, anger, sadness). However, given the continuous and complex nature of human emotions, such approaches have limitations in practical applications. In contrast, continuous emotion recognition (CER) based on the valence-arousal (VA) space offers a more nuanced representation of emotional states employing two continuous variables~\cite{Russell80}. Within this framework, valence (V) describes the positivity or negativity of an emotion, while arousal (A) quantifies its intensity (e.g., calm versus excited). This methodology has demonstrated significant value in various application domains, including intelligent human-computer interaction, mental health monitoring, autonomous driving, and intelligent education~\cite{Valstar16}.

Despite the remarkable progress achieved through deep learning techniques in affective computing, accurately modeling the temporal evolution of emotional states remains a significant challenge due to the strong temporal dependencies and complex dynamic variations inherent in emotional signals.

Affective Behavior Analysis in-the-wild (ABAW) Workshop~\cite{kollias2019deep,kollias2019expression,kollias2019face,kollias2020analysing,kollias2021affect,kollias2021analysing,kollias2021distribution,kollias2022abaw,kollias2023abaw,kollias2023abaw2,kollias2023multi,kollias20246th,kollias20247th,kollias2024distribution}will spotlight cutting-edge advancements in analyzing, generating, modeling, and understanding human affect and behavior across multiple modalities, including facial expressions, body movements, gestures and speech. A special emphasis is placed on the integration of state-of-the-art systems designed for in-the-wild analysis, enabling research and applications in unconstrained environments.

The 8th ABAW~\cite{Kollias2025,kolliasadvancements} Valence-Arousal (VA) Estimation Challenge.This challenge focuses on predicting two continuous affect dimensions —valence (ranging from -1 to 1, representing positivity to negativity) and arousal (ranging from -1 to 1,representing activity levels from passive to active). This challenge utilizes an augmented version of the audiovisual (A/V) and in-the-wild Aff-Wild2~\cite{Kollias23,kollias2023abaw,kollias2021affect} database, which includes 594 videos with approximately 3 million frames from 584 subjects.

Deep learning-based valence-arousal (VA) estimation faces several key challenges.Long-term temporal dependency: Traditional recurrent neural networks (RNNs) and long short-term memory (LSTM) networks suffer from the vanishing gradient problem when modeling long-range emotional dependencies, making it difficult to effectively capture extended emotional dynamics~\cite{Hochreiter97}.Balancing local and global temporal modeling: While convolutional neural networks (CNNs) and temporal convolutional networks (TCNs) effectively capture short-term dynamic features, they lack the capability to model global temporal information, thereby limiting the robustness of emotion recognition~\cite{Bai18}.Trade-off between computational efficiency and accuracy: Transformer-based models and their variants have demonstrated strong performance in VA estimation tasks~\cite{Vaswani17}; however, their high computational complexity poses challenges for efficient application in long-sequence tasks.

To address these challenges, researchers have proposed various methods to enhance continuous emotion recognition performance. LSTM networks have been widely adopted for temporal emotion modeling, yet their recurrent structure results in high computational costs and difficulties in capturing long-range dependencies~\cite{Chung15}. On the other hand, Transformer models and their variants, such as TimeSformer, leverage self-attention mechanisms to enhance long-term temporal modeling but suffer from high computational overhead~\cite{Poria17}. TCN, as a convolution-based temporal modeling approach, offers superior computational efficiency compared to RNNs and LSTMs; however, its ability to model long-range dependencies remains insufficient~\cite{Oord16}.

Recently, Mamba, a novel state space model (SSM), has demonstrated superior performance in long-sequence modeling tasks~\cite{Gu23}. Compared to Transformers, Mamba exhibits higher computational efficiency, and compared to RNNs, it provides more stable gradient propagation, making it well-suited for long-sequence modeling. Therefore, we hypothesize that integrating the Mamba architecture can effectively enhance the temporal modeling capability of VA-based emotion recognition.To address the aforementioned challenges, this paper proposes Mamba-VA, a novel continuous emotion recognition method based on the Mamba architecture. The key contributions of this work are as follows:

(1) Extract of high-dimensional visual features: A masked autoencoder (MAE) is employed to extract high-level visual representations from video frames, enhancing the model’s ability to encode video data.

(2) Efficient and stable temporal modeling: A TCN is utilized for short-term temporal modeling, while Mamba is employed for long-term dependency modeling, ensuring both computational efficiency and stability.

(3) Extensive evaluation on Aff-Wild2: Comprehensive experiments on the Aff-Wild2 dataset validate the strong generalization capability of Mamba-VA in continuous emotion

\section{Related Work}
\label{sec:formatting}

In previous ABAW VA estimation challenges, several research teams have proposed innovative approaches to enhance emotion recognition performance.Netease Fuxi AI Lab~\cite{zhang2024affective} developed a Transformer-based feature fusion module to comprehensively integrate emotional information from audio signals, visual images, and text, thereby providing high-quality facial features for downstream tasks. To achieve high-quality facial feature representations, they employed a masked autoencoder (MAE) as the visual feature extraction model and fine-tuned it using facial datasets. Considering the complexity of video capture scenarios, they further refined the dataset based on scene characteristics and trained classifiers tailored to each scene.

RJCMA~\cite{ling2024boosting} computed attention weights based on the cross-correlation between multimodal joint feature representations and individual modality feature representations, aiming to capture both intra-modal and inter-modal relationships simultaneously. In the recurrent mechanism, individual modality representations were reintroduced into the fusion model as input to obtain more fine-grained feature representations.

CtyunAI~\cite{dresvyanskiy2024sun} pre-trained a masked autoencoder (MAE) on facial datasets and fine-tuned it on the Aff-Wild2 dataset annotated with expression (Expr) labels. Additionally, they integrated TCN and Transformer encoders into the framework to enhance emotion recognition performance.

SUN-CE~\cite{praveen2024recursive} explored the effectiveness of fine-tuned convolutional neural networks (CNNs) and the public dimensional emotion model (PDEM) for video and audio modalities. They also compared different temporal modeling and fusion strategies by embedding these modality-specific deep neural networks (DNNs) within multi-level training frameworks.

USTC-IAT-United~\cite{yu2024multimodal} proposed the LA-SE module to better capture local image details while enhancing channel selection and suppression capabilities. They also employed a multimodal data fusion approach, integrating pre-trained audio and video backbones for feature extraction, followed by TCN-based spatiotemporal encoding and Transformer-based spatiotemporal feature modeling.

Inspired by prior research, we propose a novel emotion recognition framework that leverages MAE for high-dimensional visual feature extraction from video frames. Furthermore, we design a model comprising a four-layer TCN and Mamba architecture to enhance the effectiveness of emotion recognition in continuous valence-arousal estimation.
\section{Method}

In this section, we will provide a detailed introduction to the proposed method for the Valence-Arousal Estimation challenge in the 8th ABAW competition.
% % 插入图片代码（放在需要显示的位置）
% \begin{figure}[htbp]
%   \centering
%   \includegraphics[width=0.8\textwidth]{/image/model.png}  % 图片路径
%   \caption{The overall pipeline of our proposed method. (a) Input preprocessing; (b) Feature extraction; (c) Multi-modal fusion; (d) Regression output.}  % 标题
%   \label{fig:method_pipeline}  % 标签
% \end{figure}
\subsection{Visual Feature Extraction}

Inspired by prior works, we employ a Masked Autoencoder (MAE) as the visual feature extractor for video frames to leverage its generalizable visual representations learned through large-scale unsupervised pre-training. Unlike conventional approaches, we initialize the model with ViT-Large pre-trained weights to enhance feature representation capability and implement a parameter freezing strategy to improve generalization performance under limited labeled data.

Specifically, we freeze the parameters of the Patch Embedding layer and the first 16 Transformer Blocks while only fine-tuning the higher-level Transformer Blocks. Since the MAE has acquired rich low-level visual primitives (e.g., edges, textures, and local structures) in its early layers through large-scale unsupervised pre-training, this partial freezing strategy significantly reduces the number of trainable parameters compared to full-network fine-tuning. Such design enables more stable convergence and lower overfitting risks in data-scarce scenarios. For emotion recognition tasks requiring adaptation to facial expressions, body poses, and temporal variations, updating higher Transformer Blocks allows the model to learn task-discriminative high-level features. The CLS token embeddings, serving as global visual representations, are subsequently fed into temporal modeling modules for capturing dynamic information.

\subsection{Temporal Convolutional Network}

Videos are first split into segments with a window size $w$ and stride $s$.Given the segment window $w$ and stride $s$, a video with $n$ frames would be split into $\lfloor n/s \rfloor + 1$ segments, where the $i$-th segment contains frames $\{ F_{(i-1)*s+1}, \dots, F_{(i-1)*s+w} \}$.

The video is segmented into a series of overlapping clips, each containing a fixed number of consecutive frames. This processing approach aims to decompose the original video into smaller temporal units for efficient computational processing and analytical tasks. Crucially, the overlapping intervals between adjacent clips (typically defined by frame offsets) ensure continuous temporal coverage, thereby preserving the integrity of the visual data throughout the entire sequence.

We denote visual features as $\mathbf{f}_i$ corresponding to the $i$-th segment extracted by fine-tuned ViT-Large encoder.

Visual feature is fed into a dedicated Temporal Convolutional Network (TCN) for temporal encoding, which can be formulated as follows:

\begin{equation}
\mathbf{g}_i = \mathrm{TCN}(\mathbf{f}_i)
\end{equation}

The TCN with four hierarchical layers is employed for temporal modeling of visual features. In this architecture, the convolutional neural network processes input visual feature vectors through a sequence of convolutional layers characterized by varying kernel sizes and dilation rates. This multi-scale convolution operation generates output feature vectors with distinct dimensionality from the input, achieving feature space compression while effectively aggregating temporal contextual information across different receptive fields.

\subsection{Mamba Encoder}

In the task of affective computing, effectively modeling temporal information is crucial for accurately predicting Valence-Arousal (VA) values. To achieve this, we incorporate the Mamba module for sequential modeling, building upon the temporal features extracted by the TCN. This process can be formally expressed as:

\begin{equation}
\mathbf{m}_i = \mathrm{Mamba}(\mathbf{g}_i)
\end{equation}

Mamba is an efficient sequence modeling architecture based on the State Space Model (SSM). Compared to traditional Transformer architectures, Mamba offers lower computational complexity while demonstrating superior performance in long-sequence modeling.

After being processed by the TCN, the global interaction features $\mathbf{g}_i$ are fed into the Mamba Encoder,where they first undergo dimensional reorganization to align the channel and sequence dimensions,ensuring compatibility with standard sequential modeling formats.Subsequently,within the four-layer cascaded MambaBlock,the features at each time step activate a selective state-space mechanism—dynamically generating a state transition matrix based on the current input.This process leverages a hardware-optimized parallel scan algorithm to efficiently model long-range dependencies,while local convolutional kernels capture transient patterns in neighboring time points.

Furthermore,residual connections and layer normalization ensure stable gradient propagation.Finally,a linear projection maps the latent states into a bimodal affective space, yielding a continuous emotion trajectory.By employing an input-dependent parameter selection mechanism,Mamba adaptively focuses on emotion-relevant segments,while its linear computational complexity allows it to maintain global awareness over long sequences while enabling fine-grained analysis of subtle affective fluctuations.
\section{Experiments and Results}

\subsection{Datasets}

This Challenge’s dataset comprises 594 videos, an expansion of the Aff-Wild2 database, annotated in terms of valence and arousal. Notably, sixteen videos feature two subjects, both of whom are annotated. In total,annotations are provided for 2,993,081 frames from 584 subjects; these annotations have been conducted by four experts using the methodology outlined in [8]. Valence and arousal values are continuous and range in [-1, 1]. 

The dataset is divided into training, validation, and testing sets, ensuring subject independence, meaning each subject appears in only one set. The splits are: 356 videos in the training set; 76 videos in the Validation set; 162 videos in the testing set. 

\subsection{Implementation Details}

All models were trained on two Nvidia GeForce GTX 4090 GPUs with each having 24GB of memory.In the hyperparameter settings, we set b the number of multi-head attention to 4, the number of Mamba layers to 4, the kernel size to 15, the output feature dimension to 256, the state dimension to 8, the convolutional channel dimension to 4, and the expansion factor to 1. Additionally, we attempted to utilize a pretrained model to enhance feature extraction capabilities, but this feature was not enabled in this experiment.

For training, we employed AdamW as the optimizer and adopted the Concordance Correlation Coefficient (CCC) loss as the objective function to more accurately measure the consistency between predicted and actual emotion values. The training process was conducted over 50 epochs, with an initial learning rate of 0.0003, and a 5-epoch warmup strategy was applied to improve training stability. We also set the weight decay to 0.001 to prevent overfitting. Furthermore, we incorporated a Dropout mechanism during training, setting its rate to 0.3 to further enhance the model’s generalization ability.

\subsection{Evaluation Metrics}

The performance metric for VA challenge is the average CCC for valence and arousal, defined as: 

\begin{equation}
\mathcal{P}_{V A}=\frac{C C C_{a}+C C C_{v}}{2}
\end{equation}

The CCC measures the agreement between two time series (e.g., annotations and predictions) while penalizing deviations in both correlation and mean squared differences. CCC values range from -1 to 1, where +1 indicates perfect agreement, 0 indicates no correlation, and -1 indicates complete discordance. A higher CCC value signifies better alignment between predictions and ground-truth annotations, and therefore high values are desired. The CCC formula is defined as follows:

\begin{equation}
C C C=\frac{2 s_{x} s_{y} \rho_{x y}}{s_{x}^{2}+s_{y}^{2}+(\overline{x}-\overline{y})^{2}}
\end{equation}

where \(\rho_{x y}\) is the Pearson correlation coefficient, \(s_{x}\) and \(s_{y}\) represent the variances of the valence / arousal annotations and the predicted values, respectively, and \(s_{x y}\) is the corresponding covariance value.

\subsection{Results}

Table ~\ref{tab:example} presents the experimental results of our proposed method on the validation set for the Valence-Arousal (VA) prediction task, where the CCC is used as the evaluation metric for both valence and arousal predictions. Specifically, Fold 0 corresponds to the results on the official validation set, while Folds 1–5 represent the results obtained through five-fold cross-validation. As shown in the results, our proposed method consistently outperforms the baseline approaches.

\begin{table}
  \centering
  \begin{tabular}{@{}lccc@{}}
    \toprule
    Fold & Valence(CCC) & Arousal(CCC) & Average(CCC) \\
    \midrule
    0     & 0.5454 & 0.3848 & 0.4651 \\
    1     & 0.5423 & 0.3612 & 0.4517 \\
    2     & 0.5362 & 0.4310 & 0.4836 \\
    3     & 0.5231 & 0.3413 & 0.4322 \\
    4     & 0.5216 & 0.3540 & 0.4378 \\
    5     & 0.5305 & 0.4275 & 0.4790 \\
    \specialrule{0.6pt}{1pt}{1pt}
    Baseline & 0.2400 & 0.2000 & 0.2200 \\
    \bottomrule
  \end{tabular}
  \caption{Val Set Result}
  \label{tab:example}
\end{table}

\section{Conclusion}

This paper proposes Mamba-VA, a novel continuous emotion recognition model based on Mamba, designed to effectively model temporal emotional variations in video sequences. Our approach first employs a Masked Autoencoder (MAE) to extract high-dimensional visual features from video frames, enhancing the model’s representation capability for video data. Then, a Temporal Convolutional Network (TCN) is utilized for sequential modeling to capture local temporal dependencies, followed by Mamba for long-sequence modeling, allowing the model to learn global emotional trends effectively. Experimental results demonstrate that Mamba-VA outperforms baseline methods in the 8th ABAW Valence-Arousal Estimation Challenge, verifying its superiority in long-term emotion modeling tasks.

Compared to traditional methods, Mamba-VA integrates the advantages of CNN, TCN, and Mamba, achieving an efficient balance between computational efficiency and the ability to capture long-range dependencies in emotional states. The results on the Aff-Wild2 dataset confirm the strong generalization capability of our model, enabling stable emotion prediction in complex real-world scenarios. Additionally, our research further validates the potential of Mamba in long-sequence modeling tasks, providing an efficient and robust solution for continuous emotion recognition.

In the future, we plan to further optimize Mamba-VA by exploring multimodal fusion (e.g., audio, text, and physiological signals) to enhance emotion recognition performance. Additionally, we aim to refine training strategies to further improve the robustness and generalization ability of the model, making it more applicable to intelligent human-computer interaction, mental health monitoring, autonomous driving, and other real-world applications.

\subsection*{Acknowledgements}

This work was supported by the National Natural Science Foundation of China (Grant No.62177029)
{
    \small
    \bibliographystyle{ieeenat_fullname}
    \bibliography{main}
}


\end{document}